Data and text mining

# OpenBioLink: A benchmarking framework for large-scale biomedical link prediction

Anna Breit [1], Simon Ott [1], Asan Agibetov [1] and Matthias Samwald [1,*]

[1] Section for Artificial Intelligence and Decision Support, Medical University of Vienna, Vienna, 1090, Austria

*To whom correspondence should be addressed.

## Abstract

**Summary:** Recently, novel machine-learning algorithms have shown potential for predicting undiscovered links in biomedical knowledge networks. However, dedicated benchmarks for measuring algorithmic progress have not yet emerged. With OpenBioLink, we introduce a large-scale, high-quality and highly challenging biomedical link prediction benchmark to transparently and reproducibly evaluate such algorithms. Furthermore, we present preliminary baseline evaluation results.
**Availability and Implementation:** Source code and data are openly available at
https://github.com/OpenBioLink/OpenBioLink
**Contact:** matthias.samwald@meduniwien.ac.at

## 1 Introduction

Advances in deep learning and vector-space embedding models have enabled the creation of a sizeable array of novel methodologies for link prediction – the task of predicting missing links in knowledge graphs. As many fundamental biomedical problems can be formulated as link prediction problems, there is growing interest in the application of these algorithms in the domain of biomedicine.

Advances in methodology are both measured and steered by established general-domain benchmarks, such as the FB15K benchmark derived from Freebase, the WN18 benchmark derived from WordNet (Bordes *et al.* (2013)) or the Unified Medical Language System (UMLS) benchmark (Dettmers *et al.* (2018)).

Unfortunately, these benchmarks are often found to have flaws such as information leakage between train- and test-sets (Toutanova and Chen (2015)) and do not reflect the domain-specific properties of heterogeneous biomedical knowledge bases. Instead of capturing primarily knowledge networks (FB15K) or hierarchical taxonomies (WN18, UMLS), biomedical knowledge bases often combine richly structured ontological hierarchies with large interaction networks. Predictions of interest usually cannot be made based on simple, crisp rules. Finally, biomedical knowledge graphs tend to be large, calling into doubt whether results from smaller benchmarks are informative.

It is not straightforward to adopt existing biomedical knowledge graphs such as Bio2RDF (Dumontier *et al.* (2014)) as benchmark datasets, since they contain a significant number of metadata relations that can interfere with the performance of link prediction algorithms, and special care needs to be taken to exclude trivially inferable statements from the test set.

The first major work on embedding-based link prediction in the biomedical domain was published by Alshahrani *et al.* (2017). They evaluated a modified version of the DeepWalk algorithm adapted to heterogeneous graphs on a large-scale biomedical graph, containing Linked Data, biomedical ontologies and ontology-based annotations. Unfortunately, no benchmark dataset was established.

Crichton *et al.* (2017) and Yue *et al.* (2019) performed multiple evaluations on different graph embedding methods for link prediction, including different data sets and different train-test-set splitting techniques. This work did not focus on evaluation of heterogeneous, multi-relational graph data and corresponding algorithms. Recently PyKEEN (Ali *et al.* (2019)), a python library for training and evaluation of link prediction methods, was introduced. It offers an excellent unified interface for various graph embedding models, but no dedicated benchmark dataset was established.

A dedicated, high-quality and highly challenging benchmark optimized for the task of evaluating link prediction methods in large, heterogeneous biomedical knowledge bases has not yet been established. In this paper, we introduce the OpenBioLink suite of software, datasets and benchmarks to close this gap, and to provide a highly transparent, reproducible and configurable evaluation framework.

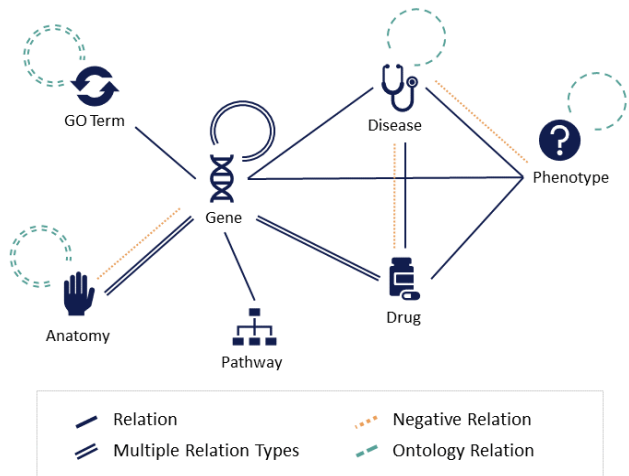

**Fig. 1.** An overview of the OpenBioLink benchmark graph.

## 2 Software Architecture

The OpenBioLink framework consists of three modules: (1) graph creation module, (2) train-test-split creation module and (3) training and evaluation module. The user can interact with these modules individually or use them together in a pipeline. The graph creation module creates the benchmark dataset from multiple public data sources. It allows for the creation of subsets and variations of the original benchmark, including the adaptation of the directionality and quality cut-off of edges as well as the exclusion of source databases or edge types. The train-test-split module divides data into a training and a test set, either randomly or via time slices. Special focus was put on the robustness and difficulty of the test set, which contains only entities that are also present in the training set and does not contain relations that can be trivially inferred from the training set (e.g., reverse edges of symmetric relations, inverse relations or super-relations). Negative samples are produced using the negative edges present in the benchmark data sources and - where needed - by applying typed negative sampling. In the third module, a model can be trained and tested. Models can be trained with external graph embedding libraries. Currently, an interface for PyKEEN is available. For evaluation, a wide range of metrics is offered, such as hits@k, mean reciprocal rank (MRR), area under the receiver operator characteristic curve (ROC AUC) and area under the precision-recall curve (PR AUC).

The OpenBioLink benchmark dataset consists of 7 node and 30 edge types, covering a wide range of ontology terms, biomedical entities and their relationships (Fig. 1). Corresponding true negative edge types used in the data set were either extracted directly from the data source or inferred from disjoint relation type pairs (e.g., for gene-anatomy relationships, over-expression and under-expression data). Statistics about the dataset are available in the supplementary material. The benchmark dataset is available in four different quality filter settings (high, medium, low and all) which are based on confidence scores. These confidence scores are data source specific, corresponding thresholds for the different quality settings are taken from the documentation of the data sources. To be applicable to a wider variety of link prediction methods, the OpenBioLink benchmark graph is available in both a directed and an undirected version. In the undirected version each relationship is present only once in the data set, while in the directed version additional explicit reverse edges for symmetric relations (e.g., "interaction") are added. Licensing terms of integrated datasets are detailed in the documentation, and should be taken into consideration when redistributing the benchmark or any derivative work.

## 3 Discussion and Future Work

A preliminary baseline evaluation with the graph embedding methods TransE (Bordes *et al.* (2013)) and TransR (Lin *et al.* (2015)) was performed. Hyperparameter optimization was performed for each model and the best model configuration was trained and tested against the OpenBioLink benchmark dataset. Details on hyperparameter estimation and per-relation results are available in supplementary table S4 and S5. The best results with hits@10 of 7.5% over all relations was achieved by a TransE model with an embedding dimensionality of 100. This result reflects that established, simple graph embedding models can make some useful predictions on this benchmark, but there is still ample room for algorithmic improvement.

We will carry out more extensive evaluations, including other methods such as the metapath-based approach (Himmelstein and Baranzini (2015)) and scalable rule-learning (Meilicke *et al.* (2019)).

To further establish the OpenBioLink framework, we will host annual, public OpenBioLink benchmarking events so that a wide range of current and upcoming link prediction models can be evaluated, and the resources of the broader research community around link prediction can be better utilized for biomedical use-cases. Future iterations of the benchmark dataset will be extended with additional knowledge from external resources such as Hetionet (Himmelstein and Baranzini (2015)).

Eventually, predictions should be verified through experiments. Ultimately, they might help improve the generation of novel research hypotheses and become an important tool for driving the advancement of biomedical research.

## Funding

This project has received funding from the European Union's Horizon 2020 research and Innovation programme under grant agreement No 668353.